\documentclass[letterpaper,10pt,conference,twocolumn]{ieeeconf} 
\IEEEoverridecommandlockouts

\usepackage{verbatim}
\usepackage[pdftex]{graphicx}
\usepackage{epsfig,subfigure}
\usepackage{amsmath,amssymb,amsfonts,bm,amscd} 
\usepackage{mathrsfs}
\usepackage{setspace}
\usepackage{fancyhdr}
\usepackage{algorithm} 
\usepackage{psfrag}
\usepackage{cite}

\usepackage{hyperref}

\usepackage{fancyhdr}
\fancypagestyle{specialfooter}{%
\fancyhf{}
\fancyfoot[L]{Some test text}
\fancyfoot[C]{\thepage}
}

\newenvironment{copyrightnoticeFont}{\fontsize{7pt}{8pt}\selectfont\fontfamily{phv}\selectfont}{\par}

\title{\LARGE \bf
Design of Deep Neural Networks as Add-on Blocks for Improving Impromptu Trajectory Tracking
}

\author{Siqi Zhou, Mohamed K. Helwa, and Angela P. Schoellig
\thanks{The authors are with the Dynamic Systems Lab (\href{http://www.dynsyslab.org}{http://www.dynsyslab.org}), Institute for Aerospace Studies, University of Toronto, Canada. M. K. Helwa is also with the Electrical Power and Machines Department, Cairo University, Egypt. This research was supported by
OCE/SOSCIP TalentEdge grant \#27901 and NSERC grant RGPIN-2014-04634. 
Emails: siqi.zhou@robotics.utias.utoronto.ca, mohamed.helwa@robotics.utias.utoronto.ca,  schoellig@utias.utoronto.ca}%
}

\makeatletter
\def\namedlabel#1#2{\begingroup
    #2%
    \def\@currentlabel{#2}%
    \phantomsection\label{#1}\endgroup
}
\makeatother

\usepackage{xcolor,colortbl}
\definecolor{grey1}{RGB}{192,192,192}
\definecolor{grey2}{RGB}{178,178,178}
\definecolor{grey3}{RGB}{150,150,150}
\definecolor{grey4}{RGB}{119,119,119}
\definecolor{grey5}{RGB}{77,77,77}
\definecolor{green}{RGB}{112,173,71}
\definecolor{blue}{RGB}{68,115,196}
\definecolor{red}{RGB}{192,0,0}
\definecolor{yellow}{RGB}{255,192,0}
\definecolor{orange}{RGB}{237,125,49}

\newcommand{\com}{\textcolor{black}}
\newcommand{\edit}{\textcolor{black}}
\newcommand{\rev}{\textcolor{black}}

\usepackage[colorinlistoftodos]{todonotes}

\usepackage{bm}

\begin{document}

\maketitle
\thispagestyle{fancyplain}
\renewcommand{\headrulewidth}{0pt}
\pagenumbering{gobble}
 \lfoot{\begin{copyrightnoticeFont}\vspace{-2em}
 \textbf{Accepted final version}. To appear in \textit{the 56th IEEE Conference on Decision and Control} (CDC 2017).\\
 \copyright2017 IEEE. Personal use of this material is permitted. Permission from IEEE must be obtained for all other uses, in any current or future media, including reprinting/republishing this material for advertising or promotional purposes, creating new collective works, for resale or redistribution to servers or lists, or reuse of any copyrighted component of this work in other works.\end{copyrightnoticeFont}}
\pagestyle{empty}

\begin{abstract}
This paper introduces deep neural networks (DNNs) as add-on blocks to baseline feedback control systems to enhance tracking performance of arbitrary desired trajectories. The DNNs are trained to adapt the reference signals to the feedback control loop. The goal is to achieve a unity map between the desired and the actual outputs. In previous work, the efficacy of this approach was demonstrated on quadrotors; on 30 unseen test trajectories, the proposed DNN approach achieved an average impromptu tracking error reduction of 43\% as compared to the baseline feedback controller. Motivated by these results, this work aims to provide platform-independent design guidelines for the proposed DNN-enhanced control architecture. In particular, we provide specific guidelines for the DNN feature selection, derive conditions for when the proposed approach is effective, and show in which cases the training efficiency can be further increased.
\end{abstract}


\setcounter{section}{0}
\section{Introduction}
\label{sec:introduction}

High-accuracy trajectory tracking is an essential requirement for many industrial applications including robot-aided inspection and advanced manufacturing~\cite{nikolic2013uav,su2004disturbance}. 
\com{Classical control methods for trajectory tracking, such as model predictive control (MPC) or PID control, either require a sufficiently accurate dynamic model of the plant~\cite{rawlingandmayne2015}, which may not be available, or rely on manual tuning of the system controller parameters, which can be difficult and time-consuming, and may result in overly conservative behavior~\cite{aastrom2004revisiting}}.
Learning methods such as iterative learning control (ILC) have been successfully applied to many robotic applications to improve the tracking performance through repeated trials \cite{schoellig,ostafew2013visual}. However, this requires repeated training of the known, desired trajectory. In this paper, we consider a more challenging problem: impromptu tracking; that is, to accurately track an arbitrary reference in a single attempt.

\begin{figure}[t]
\vspace{0.8em}
\centering
\includegraphics[width=0.475\textwidth]{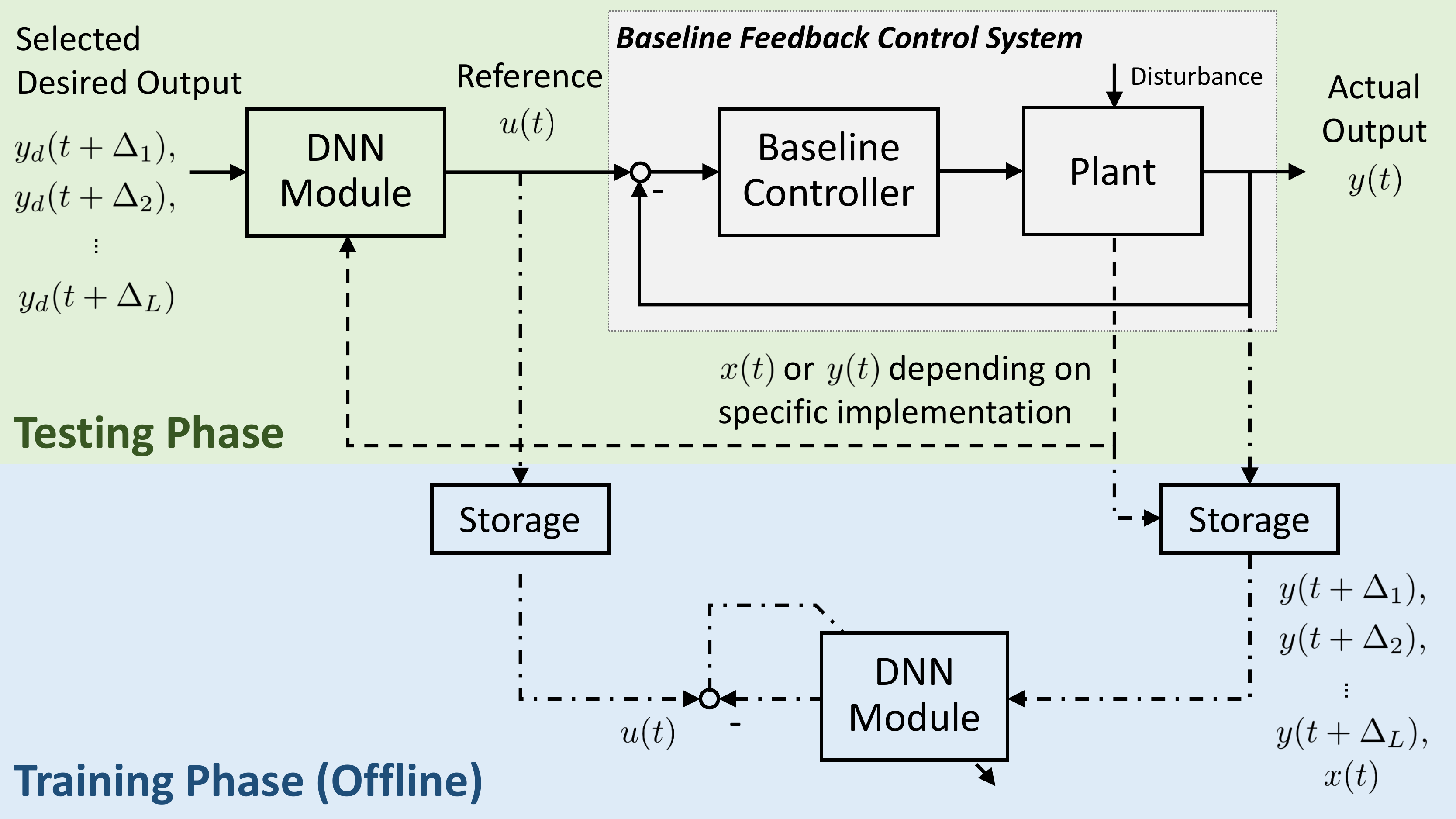}
\caption{An illustration of the proposed control architecture and training process considered in this work. During the testing phase, at a particular time step $t$, the selected desired output \{$y_{d}(t+\Delta_1)$, $y_{d}(t+\Delta_2)$,..., $y_{d}(t+\Delta_L)$\} and the current state $x(t)$ (or the current output $y(t)$) are inputs to the DNN module to generate reference signals $u(t)$ for the baseline system, where $t\in \mathbb{Z}_{\ge 0}$ is the discrete-time index and $\Delta_i \in \mathbb{Z}_{\ge 0}$ for all $i$. During the training phase, a set of training trajectories are performed on the baseline system, and $u(t)$, $y(t)$, and $x(t)$ are stored for training the DNNs offline.}
\label{fig:blkdiag}
\vspace{-1em}
\end{figure}

Given the ability of deep neural networks (DNNs) to generalize knowledge, a DNN-enhanced control architecture has been proposed in~\cite{DNNimpromptuTrack} to improve the tracking performance of traditional feedback controllers for any given desired trajectory.
In this proposed architecture (illustrated in Fig.~\ref{fig:blkdiag}), a DNN module is pre-cascaded to a baseline feedback control system to adjust the reference inputs to the feedback control system with the goal of achieving perfect output tracking. On 30 test trajectories, the DNN-based approach led to an average of 43\% tracking error reduction as compared to the baseline controller~\cite{DNNimpromptuTrack}.
Though the effectiveness of the architecture was well-illustrated on quadrotor vehicles, general DNN design guidelines were not provided in~\cite{DNNimpromptuTrack}. In particular, the inputs and outputs of the DNN module were determined through experimental trial-and-error. 
Similar to the fundamental work on feature selection for image classification and speech recognition, in this work we derive rules for feature selection in the context of tracking control. 
\rev{In addition, we derive conditions under which the proposed approach is effective and identify cases for which the training efficiency can be further improved. These contributions together provide a platform-independent formulation of the approach utilized in~\cite{DNNimpromptuTrack}.} 

Below we first provide a brief review of the relevant literature (Section~\ref{sec:relatedWork}). We then define our problem (Section~\ref{sec:problemStatement}), derive theoretical insights on the DNN module design (Section~\ref{sec:results}), and present simulation and experimental results (Sections~\ref{sec:simulation} and \ref{sec:experiment}).

\section{Related Work}
\label{sec:relatedWork}
In the literature, there are many examples where machine learning techniques have been successfully integrated with control system designs to improve the tracking performance in uncertain environments~\cite{berkenkamp2015safe,bristow2006survey,assael2015data,schoellig}. As discussed in~\cite{DNNimpromptuTrack}, examples of these machine learning techniques include, but are not limited to, Gaussian Processes (GPs), ILC, and DNNs. 
In comparison, DNNs are flexible modeling frameworks capable of approximating highly nonlinear functions and generalizing learned experience to unseen situations~\cite{hunt1992neural}. 
As compared to GPs specifically, DNNs have the advantage that their computation time and memory requirement do not increase with the size of the training dataset~\cite{DNNimpromptuTrack}; 
as compared to ILC, DNN-based approaches can be more conveniently generalized to untrained tasks~\cite{bristow2006survey}. 


\edit{As summarized in early review papers \cite{hunt1992neural,balakrishnan1996neurocontrol}, DNNs have been combined with various control techniques such as \textit{predictive control} and \textit{adaptive control} to directly or indirectly account for uncertainties and non-idealities in the overall system.} 
In the recent literature, there are many advanced works illustrating the efficacy of utilizing DNNs in different control applications. For instance, in~\cite{Bansal}, it is illustrated that DNNs can approximate the unmodeled dynamics of quadrotors to assist linear quadratic regulators (LQR) in trajectory tracking control. 
\com{In~\cite{he2016adaptive} and the references therein, DNNs are used to compensate for the model uncertainties in impedance control of manipulators.} 
\edit{Moreover, in~\cite{assael2015data}, for single and double pendulums, \rev{DNN models are utilized in nonlinear model predictive control (NMPC) for determining the optimal control input.} In~\cite{sanchez2016real}, simulations on several aerospace systems illustrate that DNNs can also be used to approximate solutions of the Hamilton-Jacobi-Bellman (HJB) equations to save the online computation time.} 

Adding to this body of work, \cite{DNNimpromptuTrack} proposes pre-cascading a DNN module 
to a feedback control system of a quadrotor to achieve a unity mapping from the desired output to the actual output. By aiming for a unity map, this methodology is similar to \textit{direct inverse control} and \textit{adaptive inverse control}~\cite{frye2014direct,hunt1992neural,ge2003neural,zhang2016adaptive,chen1995adaptive}. However, there are several major differences. One of the major differences is that the DNN models in~\cite{DNNimpromptuTrack} are not directly applied to the open-loop plant. Instead, they modify the reference of a stable closed-loop system and can run at a lower rate, as compared to the underlying baseline controller, which makes the approach less prone to stability issues. Moreover, in adaptive inverse control, the weights of the DNN models must be updated online to ensure the stability of the overall system~\cite{ge2003neural}, and the convergence of the weights requires good initializations~\cite{chen1995adaptive}. In contrast, with the proposed architecture in~\cite{DNNimpromptuTrack}, improving tracking performance is decoupled from achieving stability. Stability is guaranteed by appropriately designing the controller of the baseline closed-loop system.
Improvements to the tracking performance are achieved by the DNNs. 
\rev{Furthermore, as opposed to~\cite{zhang2016adaptive}, where recurrent neural networks (RNNs) are used, in the proposed architecture, feedforward neural networks (FNNs) are used for model learning. This makes the proposed architecture less prone to instability issues and simplifies practical implementations~\cite{jaeger2002tutorial}.} 
Given these advantages and further motivated by the successful results in~\cite{DNNimpromptuTrack}, in this work, we aim to further analyze the DNN-enhanced control architecture of~\cite{DNNimpromptuTrack}, and identify conditions where the proposed approach is most effective and efficient.
\section{Problem Statement}
\label{sec:problemStatement}
We consider the control architecture shown in Fig.~\ref{fig:blkdiag}. A DNN is introduced as an add-on module to a closed-loop, stable system in order to improve the tracking control performance. We aim to derive general guidelines for the design of the DNN module. This includes:
\begin{enumerate}
\item identification of conditions under which the add-on DNN module improves the system's tracking performance,
\item rules for the DNN feature selection, and
\item characterization of conditions under which the efficiency of the training can be further improved.
\end{enumerate}
In order to address the aspects outlined above, we first assume that the underlying closed-loop system can be described by a linear time-invariant (LTI), single-input-single-output (SISO) system. This discussion is then extended to nonlinear systems. In particular, we assume the discrete-time dynamics of the baseline system can be represented by
\begin{equation}
\begin{aligned}
\label{eqn:system_linear}
x(t+1) = Ax(t)+bu(t),\:\:y(t)= cx(t)
\end{aligned}
\end{equation}
in the linear scenario, and by
\begin{equation}
\begin{aligned}
\label{eqn:system_nonlinear}
x(t+1) = f\left(x(t)\right)+g\left(x(t)\right)u(t),\:\:
y(t)= h\left(x(t)\right)
\end{aligned}
\end{equation}
in the nonlinear scenario, where $t\in \mathbb{Z}_{\ge 0}$ is the discrete-time index, $x\in\mathbb{R}^n$ is the state, $y\in\mathbb{R}$ is the output, $u\in\mathbb{R}$ is the reference, $A$, $b$, and $c$ are constant matrices of appropriate dimensions, and $f(\cdot)$, $g(\cdot)$, and $h(\cdot)$ are smooth functions.





\section{Main Results}
\label{sec:results}
In this section, given the system representations (\ref{eqn:system_linear}) and (\ref{eqn:system_nonlinear}), we first identify the function that the DNN would need to represent. We then use this result to derive \textit{(i)}~necessary conditions for the add-on DNN module to be effective (Section~\ref{subsec:zeroDynamics}) and \textit{(ii)}~the DNN input features necessary to model the underlying function (Section~\ref{subsec:featureSelections}). Moreover, we derive necessary conditions enabling further improvements of the
efficiency of the DNN training (Section~\ref{subsec:differenceLearning}). Even though these discussions start from known dynamics of the baseline feedback loop, in practice only minimal knowledge of the closed-loop system is needed. As will be discussed in detail, this knowledge can be either obtained from simple identification experiments such as step response tests or from basic knowledge about the system. 

\subsection{Underlying Function Modeled by the DNNs}
\label{subsec:zeroDynamics}
The DNN add-on module aims to establish an identity mapping from the desired output $y_d$ to the actual output $y$. In this part, we will show that the function approximated by the DNN module is the output equation of the inverse dynamics of the feedback control loop. Stable zero dynamics of the baseline system is consequently a necessary condition for the proposed approach to be effective. 

For the convenience of discussion, we first state the definition of the relative degree of a dynamical system. For the linear SISO system~(\ref{eqn:system_linear}), the \textit{relative degree} is the smallest integer $r$ for which $cA^{r-1}b \neq 0$~\cite{henson1997nonlinear}. 
Using this definition, it can be shown that the input and output of (\ref{eqn:system_linear}) are related by
\begin{align}
\label{eqn:underlyingFunction_linSS_y}
y(t+r) = cA^rx(t)+cA^{r-1}bu(t).
\end{align}
Then, at time step $t$, by selecting 
\begin{equation}
\label{eqn:underlyingFunction_linSS}
u(t) = \frac{1}{cA^{r-1}b}\left(-cA^rx(t)+y_d(t+r) \right),
\end{equation}
$y(t+r)=y_d(t+r)$ is satisfied as desired. By considering $y_d(t+r)$ as the input and $u(t)$ as the output, Eqn. (\ref{eqn:underlyingFunction_linSS}) is in fact the output equation of the inverse dynamics of the baseline system~(\ref{eqn:system_linear}). Thus, recalling the architecture in Fig.~\ref{fig:blkdiag}, by training the DNN module to approximate Eqn.~(\ref{eqn:underlyingFunction_linSS}), exact tracking can be achieved in theory.

Note that there is an inherent delay of $r$ time steps from the input to the output in Eqn.~(\ref{eqn:underlyingFunction_linSS_y}), which is a known fact for discrete-time systems with relative degrees $r$. \edit{From Eqn.~\eqref{eqn:underlyingFunction_linSS}, at a particular time step $t$, the computation of $u(t)$ depends on $y_d(t+r)$ to compensate for the inherent delay. In practice, for off-line or on-line trajectory generation algorithms, a preview of $r$ steps of the desired trajectory $y_d$ is typically available; hence, the non-causality in Eqn.~\eqref{eqn:underlyingFunction_linSS} is \textit{not} an issue in practice.}

The above analysis can be extended to nonlinear systems. Following the discussion outlined in~\cite{sun2001analysis}, we use $h\circ f$ to denote the composition function of $f$ and $h$, and $f^i$ to denote the $i^{\text{th}}$ composition of the function $f$ with $f^0(x(t)) = x(t)$ and $f^i(x(t))=f^{i-1}\circ(f(x(t)))$. Around an operating point $(x_0,u_0)$, the \textit{relative degree} of system~(\ref{eqn:system_nonlinear}) is defined as the smallest integer $r$ such that $\frac{\partial}{\partial u}h \circ f^{r-1}(f(x(t))+g(x(t))u(t))\neq 0$ for each point in the neighborhood of the operating point. We assume that system~\eqref{eqn:system_nonlinear} has a well-defined relative degree $r$ in the operating region. Then, it can be shown that the input and the output of system~(\ref{eqn:system_nonlinear}) are related by
\begin{equation}
\label{eqn:underlyingFunction_nonlinSS_y}
y(t+r) = h\circ f^{r-1}\big(f(x(t))+g(x(t))u(t)\big).
\end{equation}
Assuming $y(t+r)$ is affine in $u(t)$, Eqn.~(\ref{eqn:underlyingFunction_nonlinSS_y}) can be simplified to
\begin{equation}
\label{eqn:underlyingFunction_nonlinSS_y_sim}
y(t+r)=\hat{h}(x(t)) + D(x(t))u(t),
\end{equation}
where $\hat{h}(x(t))= h\circ f^r(x(t))$ and $D(x(t))=\frac{\partial}{\partial u}h \circ f^{r-1}(f(x(t))+g(x(t))u(t))$ \cite{sun2001analysis,jang1994iterative}. Note that in Eqn.~(\ref{eqn:underlyingFunction_nonlinSS_y_sim}), $D(x(t))\neq 0$ by the definition of the relative degree. Following the same argument as for linear systems, the control law for achieving $y(t+r)=y_d(t+r)$ is 
\begin{equation}
\label{eqn:underlyingFunction_nonlinear_u}
u(t) = \frac{1}{D(x(t))}\left(-\hat{h}(x(t))+y_d(t+r) \right)
\end{equation}
for the affine case in Eqn.~(\ref{eqn:underlyingFunction_nonlinSS_y_sim}), and it is reasonable to assume that
\begin{equation}
\label{eqn:underlyingFunction_nonlinSS}
u(t) = F\left(x(t),y_d(t+r) \right)
\end{equation}
for the general case in Eqn.~(\ref{eqn:underlyingFunction_nonlinSS_y}), where $F$ is a nonlinear function to be approximated by the DNN module.

From Eqn.~(\ref{eqn:underlyingFunction_nonlinear_u}) and Eqn.~(\ref{eqn:underlyingFunction_nonlinSS}), the required information for computing $u(t)$ remains the same as that for linear systems. In particular, the function depends on the current state $x(t)$ and the desired output $r$ steps in the future $y_d(t+r)$, where $r$ is the relative degree of the baseline system.\\[0.5em]
\noindent\textbf{Insight 1a.} \textbf{Underlying Function:} In order to achieve an identity mapping from the desired output to the actual output, the DNN module should approximate the output equation of the inverse dynamics of the baseline system, which corresponds to Eqn.~(\ref{eqn:underlyingFunction_linSS}) for the linear scenario and Eqn.~(\ref{eqn:underlyingFunction_nonlinSS}) for the nonlinear scenario.

\noindent\textbf{Insight 1b.} \textbf{Necessary Conditions for the Effectiveness of the Approach:} The effectiveness of our proposed DNN architecture depends on two conditions: \textit{(i)} the baseline system has a well-defined relative degree $r$ (within the defined operating region) and \textit{(ii)} the inverse dynamics of the system are stable. \\[0.5em]
\indent \textit{As a result, for implementing the proposed DNN-enhanced architecture, we only need to identify the relative degree of the feedback system rather than its exact dynamics.} \edit{For linear and nonlinear discrete-time systems, the relative degree is the number of time steps of delay between a change in the input and the resulting change in the output, which can be easily determined from the system's step response in practice.} 
\rev{In~\cite{DNNimpromptuTrack}, it is proposed to run the DNN module at a lower sampling frequency as compared to the baseline system to avoid instability problems; in this case, relative degrees should be determined with respect to the DNN sampling frequency.} For nonlinear systems, since relative degrees are locally defined \cite{henson1997nonlinear}, having multiple DNN modules for different operating regions may be required. 

For linear systems, the stability of the inverse dynamics is equivalent to the stability of the system's zero dynamics, which is characterized by the zeros of the system's transfer function; this can be also checked experimentally using the system's step response~\cite{henson1997nonlinear}. For nonlinear systems, a necessary condition for the stability of the inverse dynamics is the stability of the system's zero dynamics, which are the \rev{invariant dynamics} of (\ref{eqn:system_nonlinear}) when the input $u(t)$ is selected such that the output $y(t)$ is forced to be $0$ at all time steps. Note that for nonlinear systems, this condition is, however, not sufficient~\cite{sussmann1990limitations,isidori2013nonlinear}.\textit{ Therefore, for both systems~(\ref{eqn:system_linear}) and~(\ref{eqn:system_nonlinear}), a necessary condition for the stability of the inverse dynamics, and hence for the effectiveness of the DNN-based approach, is the stability of the zero dynamics of the baseline system.}

\subsection{Feature Selection}
\label{subsec:featureSelections}
In this subsection, we discuss the correct selection of input features for the DNN module to achieve high tracking performance. 
Based on the state space representation in the previous section, and from Eqn.~\eqref{eqn:underlyingFunction_linSS} and Eqn.~\eqref{eqn:underlyingFunction_nonlinSS}, 
the information for the DNN module to correctly generate the reference $u(t)$ consists of the current state $x(t)$ and the future desired output $y_d(t+r)$, where $r$ is the system's relative degree. 
For linear systems, an alternative feature selection can be obtained using the transfer function of the system~(\ref{eqn:system_linear}). In particular, assuming zero initial conditions, the equivalent $z$-transform of system~(\ref{eqn:system_linear}) is
\begin{align}
\frac{Y(z)}{U(z)}
\label{eqn:system_linear_TF}
= \frac{\beta_{n-r}z^{n-r}+\beta_{n-r-1}z^{n-r-1}+\cdots +\beta_0}{z^n + \alpha_{n-1}z^{n-1}+\cdots+\alpha_0},
\end{align}
where $Y(z)$ and $U(z)$ are the $z$-transforms of the output and reference of the system, $n$ is the system's order and $r$ is its relative degree, and $\alpha_i$ and $\beta_i$ are constants. 
Assuming both the dynamics and the zero dynamics of (\ref{eqn:system_linear_TF}) are stable, it can be easily verified that by choosing the following control law
\begin{equation}
\begin{aligned}
\label{eqn:underlyingFunction_tf_u}
u(t) 
&= \frac{1}{\beta_{n-r}}y_d(t+r)+\frac{\alpha_{n-1}}{\beta_{n-r}}y_d(t+r-1)+\cdots \\
&\hspace{1em}+\frac{\alpha_0}{\beta_{n-r}} y_d(t-n+r)-\frac{\beta_{n-r-1}}{\beta_{n-r}}u(t-1)\\
&\hspace{1em}-\frac{\beta_{n-r-2}}{\beta_{n-r}}u(t-2)-\cdots-\frac{\beta_0}{\beta_{n-r}} u(t-n+r),
\end{aligned}
\end{equation}
exact tracking is achieved. Based on Eqn.~\eqref{eqn:underlyingFunction_tf_u}, the input features of the DNN can be selected as \{$y_d(t+r)$, $y_d(t+r-1)$,..., $y_d(t-n+r)$, $u(t-1)$, $u(t-2)$,..., $u(t-n+r)$\}.\\[0.5em]
\noindent\textbf{Insight 2.} \textbf{Feature Selection:} By associating the DNN module with the inverse dynamics of the baseline feedback control loop, the state space and transfer function representations provide two approaches for selecting the input features of the DNN module to achieve exact tracking. In particular, for the state space representations of linear and nonlinear systems, the input features to the DNN module are $x(t)$ and $y_d(t+r)$. For the transfer function representation of linear systems, the input features are \{$y_d(t+r)$, $y_d(t+r-1)$,..., $y_d(t-n+r)$, $u(t-1)$, $u(t-2)$,..., $u(t-n+r)$\}.



\edit{The transfer function approach does not require the knowledge of the internal state $x(t)$ of the system, but is limited to linear systems. 
In practice, this approach may be applied to cases where the \rev{baseline system can be approximated by linear dynamics (e.g.,~\cite{helwa2016construction,giesbrecht2009vision})}. 
For general nonlinear control problems, one may utilize the state space approach to implement the DNN module together with standard state estimation techniques. The inclusion of the state in the DNN inputs also allows this approach to partially compensate for initial errors or disturbances in the system. 
Furthermore, the dimension of the input to the DNN is $(n+1)$ for the state space approach and $(2n-r+1)$ for the transfer function approach. For high-dimensional systems with low relative degrees, the state space approach may be preferable because of the relatively smaller DNN input dimension.
}

\subsection{Improvement of Training Efficiency}
\label{subsec:differenceLearning}
\edit{In the implementation of~\cite{DNNimpromptuTrack}, the position components of the input and output of the DNN module were taken relative to the current position (referred to as the \textit{difference learning scheme}) to simplify the training process. 
By using relative terms, the function learned by the DNN module is translationally invariant with respect to position. Consequently, the amount of data needed for the training is reduced, and the training process becomes more efficient. In this subsection, we prove that there is an implicit assumption for the difference learning scheme to be valid. }
In particular, we consider the linear system~\eqref{eqn:system_linear} using both transfer function and state space formulations. For the state space formulation, we assume that the output $y=x_1$, the first element of the state vector $x$, and that for step inputs, the steady state values of the remaining states $x_2,\cdots,x_n$ are all zeros. This assumption is valid, for instance, for mechanical systems with position-velocity state space. 
\\[0.5em]
\noindent\rev{\noindent\textbf{Lemma 1.}\textbf{ A Necessary Condition for Difference Learning:}}
Consider system~\eqref{eqn:system_linear} and the DNN-enhanced architecture in Fig.~\ref{fig:blkdiag}. Then, the DNN-enhanced approach with a difference learning scheme is able to achieve exact tracking 
only if the baseline system has a unity DC gain. 
\\[0.5em]
%
\noindent\textit{Proof.} \rev{We first consider the transfer function formulation in Eqn.~(\ref{eqn:underlyingFunction_tf_u}), which corresponds to the exact inverse of system~\eqref{eqn:system_linear}. 
By defining $\Delta_u(t+k) := u(t+k) - y_d(t)$ and $\Delta_{y_d}(t+k):=y_d(t+k)-y_d(t)$ for $k\in \mathbb{Z}$, it can be shown that Eqn.~(\ref{eqn:underlyingFunction_tf_u}) can be rewritten as}
\begin{equation}
\begin{aligned}
\label{eqn:underlyingFunction_tf_u_diff}
\Delta_{u}(t) &= \frac{1}{\beta_{n-r}}\Delta_{y_d}(t+r)+\frac{\alpha_{n-1}}{\beta_{n-r}}\Delta_{y_d}(t+r-1)+\cdots \\&\hspace{1em}+\frac{\alpha_0}{\beta_{n-r}} \Delta_{y_d}(t-n+r)-\frac{\beta_{n-r-1}}{\beta_{n-r}}\Delta_{u}(t-1)\\
&\hspace{1em}-\frac{\beta_{n-r-2}}{\beta_{n-r}}\Delta_{u}(t-2)-\cdots-\frac{\beta_0}{\beta_{n-r}}\Delta_{u}(t-n+r)\\
&\hspace{1em}+\underbrace{\frac{1}{\beta_{n-r}}\left(1-\sum_{i=0}^{n-r}\beta_i +\sum_{i=0}^{n-1}\alpha_i \right)y_d(t)}_{\triangleq s(y_d(t))}.
\end{aligned}
\end{equation}
\edit{On the right-hand side of Eqn.~(\ref{eqn:underlyingFunction_tf_u_diff}), the only term containing the non-difference, time-dependent variable $y_d(t)$ is $s(y_d(t))$.} It is possible to express $\Delta_{u}(t)$ as a function of the remaining $\Delta_{y_d}$ and $\Delta_{u}$ terms and hence utilize the difference learning scheme if and only if $s(y_d(t))=0$. For arbitrary $y_d(t)$, this condition implies $\sum_{i=0}^{n-r}\beta_i=1+\sum_{i=0}^{n-1}\alpha_i$. 
By examining the transfer function in Eqn.~(\ref{eqn:system_linear_TF}), this is equivalent to the condition of having a baseline system with a unity DC gain. 

The same condition can be derived for the state space formulation in (\ref{eqn:underlyingFunction_linSS}). 
For this case, in order to introduce translational invariance with respect to $y_d$, the inputs to the DNN module are defined as $\Delta_{y_d}(t+r):=y_d(t+r)-y_d(t)$ and $\Delta_x(t) := x(t)-\begin{bmatrix}y_d(t)&0\cdots&0\end{bmatrix}^\intercal$, and the output is $\Delta_u(t)=u(t)-y_d(t)$. 
\edit{For proving Lemma~1 in this case, we show that if the DC gain of the feedback system~(\ref{eqn:system_linear}) is not unity, then the DNN trained based on the difference learning scheme cannot correct constant errors (offsets) and the overall DNN-enhanced system will not be able to achieve zero steady state errors for constant references.} 
Consider a baseline feedback system with a DC gain $K_0 \neq 1$. Suppose, by contradiction, that we have zero steady state error for the desired step reference $y_d(t)=K$, for all $t\geq 0$ and arbitrary constants $K$. For this case, for all $t\geq 0$, $y_d(t+r)=y_d(t)=K$ and $\Delta_{y_d}=0$ uniformly. Also, since the steady state error is zero by assumption, $\Delta_{x}=0$ at the steady state. Thus, at the steady state, the inputs to the DNN are all zeros, and $\Delta_u=\bar{b}$, where $\bar{b}$ is the network constant bias, which is preselected in the training phase. This implies that the input to the baseline feedback system at the steady state is $K+\bar{b}$, and consequently, the steady state output of the system is $y_{ss}=K_0(K+\bar{b})$. Since $y_{ss}=K$ by assumption, we have $\bar{b}=(K/K_0)-K$. Since $K_0\neq 1$ by assumption, $\bar{b}$ is dependent on the arbitrary value $K$, a contradiction to the fact that $\bar{b}$ is a constant value.\hfill$\square$

From the proof above, the correct output of the DNN module at steady state to achieve zero steady state error for a step reference $K$ is $\Delta_u=(K/K_0)-K$. Since $K$ is arbitrary, then, except for the case that $K_0=1$, the mapping from $\Delta_x(t)=0$ and $\Delta_{y_d}(t+r)=0$ to $\Delta_u(t)$ is one-to-many, which cannot be effectively learned by DNNs~\cite{jordan1992forward}. 
In the above discussion, we used $y_d(t)$ as the reference point in the difference learning scheme. In practice, during the testing phase, $y_d(t)$ on either or both input and output sides of the DNN module can be replaced with $y(t)$ to additionally compensate for initial errors and disturbances. \\[0.5em]
\noindent\textbf{Insight 3.}\textbf{ A Necessary Condition for Difference Learning:} Based on the above theoretical study for linear systems, for the difference learning scheme utilized in~\cite{DNNimpromptuTrack} to be valid, the baseline system must have a unity DC gain. If one plans to use the difference learning scheme to improve the DNN training efficiency, then the baseline system should be designed to achieve zero/small steady state errors for step references. 

In Section~\ref{sec:experiment}, we verify the necessity of this condition for nonlinear systems through experiments.

\section{Simulation Results}
\label{sec:simulation}
In order to illustrate \textit{Insight~1}, we consider two LTI, SISO systems, representing the baseline, closed-loop dynamics in Fig.~\ref{fig:blkdiag}:
\begin{equation}
\label{eqn:system_sim_stable}
\begin{aligned}
x(t+1) &= \left[\begin{matrix}
0 & 1\\-0.15 & 0.8
\end{matrix}\right]x(t)+\left[\begin{matrix}
0 \\ 1
\end{matrix}\right]u(t)\\
y(t) &= \left[\begin{matrix}
-0.2 & 1
\end{matrix}\right]x(t), 
\end{aligned}
\end{equation}
\begin{equation}
\label{eqn:system_sim_unstable}
\begin{aligned}
x(t+1) &= \left[\begin{matrix}
0 & 1\\-0.15 & 0.8
\end{matrix}\right]x(t)+\left[\begin{matrix}
0 \\ 1
\end{matrix}\right]u(t)\\
y(t) &= \left[\begin{matrix}
-450.9 & 450
\end{matrix}\right]x(t).
\end{aligned}
\end{equation}
The two systems have the same poles but different zeros.  
System~(\ref{eqn:system_sim_stable}) has a stable (minimum phase) zero at 0.2, and~(\ref{eqn:system_sim_unstable}) has an unstable (non-minimum phase) zero at 1.002.
For the purpose of illustrating \textit{Insight~1}, the state space approach is used for selecting the features of the DNN module. At a particular time step $t$, the input to the DNN module is $\mathcal{I}=\{x(t), y_d(t+r)\}$, and the output is $\mathcal{O}=\{u(t)\}$. 
The equation to be approximated by the DNN module is Eqn.~(\ref{eqn:underlyingFunction_linSS}). For this simulation study, with known system matrices $(A,b,c)$, Eqn.~(\ref{eqn:underlyingFunction_linSS}) can be computed at each time step and utilized to examine \textit{Insight~1}.

\subsection{DNN Architecture and Training}
\label{subsec:sim_DNN}
For this simulation example, Matlab's Neural Network Toolbox is used for constructing and training the DNN models. 
\edit{For both systems, the DNN models are fully-connected FNNs with 2 hidden layers of 20 hyperbolic tangent neurons. 
}
The training data is generated from the baseline system response to sinusoidal trajectories with 25~different combinations of amplitudes $\{1,2,3,4,5\}$ and frequencies $\{0.024,0.032,0.048,0.091,1.000\}~\text{Hz}$. To avoid overfitting a particular training trajectory, the training dataset is constructed using a balanced number of randomly chosen data points from each trajectory. \edit{The Levenberg-Marquardt algorithm is used for DNN model parameter training to minimize the mean squared error between the DNN's output and the target from the training dataset. For both systems, the DNN training converges within 1000 iterations.} We test the overall DNN-enhanced systems on a set of untrained trajectories.

\subsection{Testing Results} 
\label{subsec:sim_result}
In order to illustrate \textit{Insight~1}, the baseline and DNN-enhanced performance of system~(\ref{eqn:system_sim_stable}) and system~(\ref{eqn:system_sim_unstable}) are tested for the trajectory $y_d(t)=\sin\left(\frac{2\pi}{15} t\right)+\cos\left(\frac{2\pi}{12} t\right) - 1$.
\begin{figure}[!t]
\vspace{0.2em}
\centering
\subfigure[References $u$\label{subfig:sim_ref}]{\includegraphics[trim={0 0 0 0},clip,width = 0.237\textwidth]{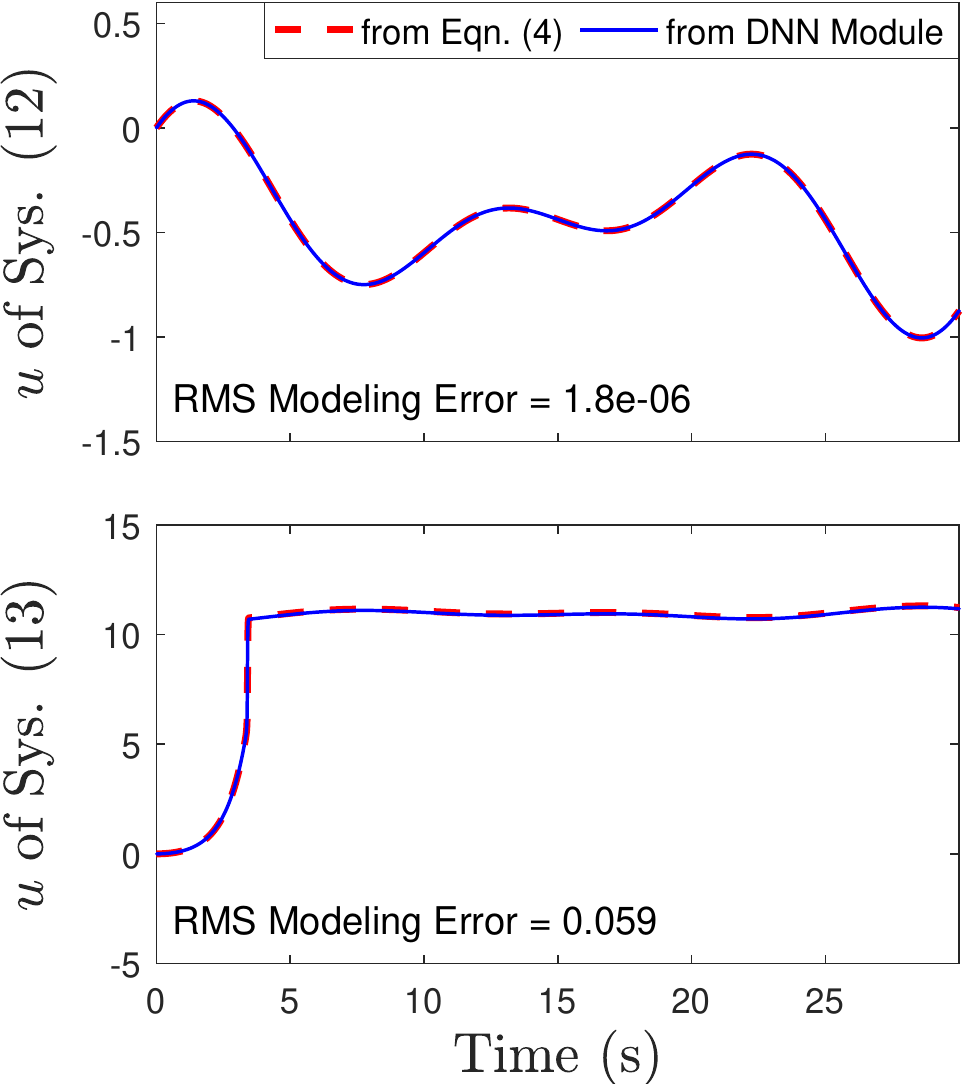}}
\subfigure[Outputs $y$\label{subfig:sim_output}]{\includegraphics[trim={0 0 0 0},clip,width = 0.237\textwidth]{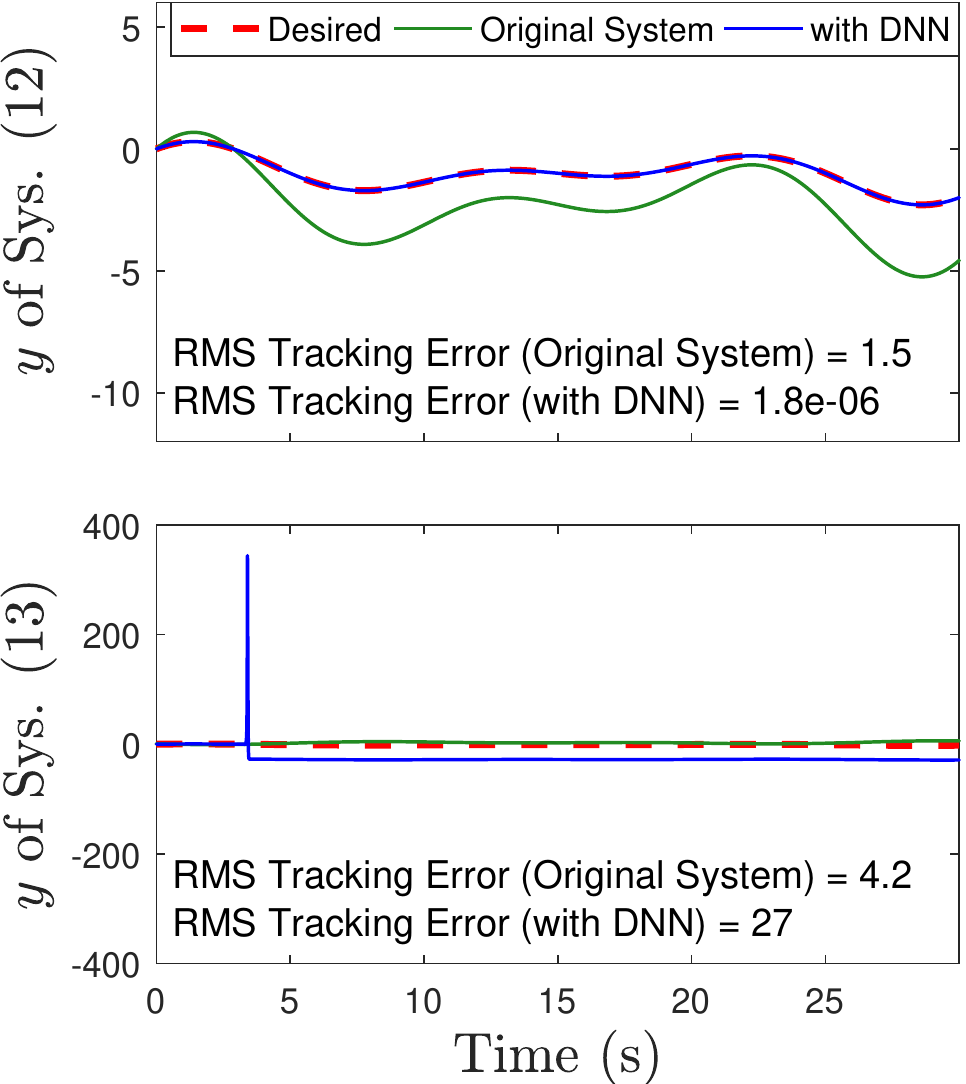}}
\vspace{-1em}
\caption{
The references and outputs of the closed-loop systems~(\ref{eqn:system_sim_stable}) and (\ref{eqn:system_sim_unstable}) for the test trajectory $y_d(t)=\sin\left(\frac{2\pi}{15} t\right)+\cos\left(\frac{2\pi}{12} t\right) - 1$. From (a), the DNN accurately learns the output equation of the inverse dynamics of the baseline systems. From (b), for system~(\ref{eqn:system_sim_stable}), the DNN-enhanced approach is able to approximately achieve exact tracking; however, for system~(\ref{eqn:system_sim_unstable}), which has an unstable zero, the inherent instability causes numerical issues and prevents the DNN-enhanced approach from being effective.
}
\label{fig:simulation}
\vspace{-1em}
\end{figure}
\com{From Fig.~\ref{subfig:sim_ref}, it can be seen that for both systems, 
the reference signal outputted by the DNN module (blue) and that computed based on Eqn.~(\ref{eqn:underlyingFunction_linSS}) (red) almost coincide.} This verifies our theoretical insight that the underlying function learned by the DNN module is the output equation of the inverse dynamics of the baseline feedback control system. This also shows that a well-trained DNN module can accurately approximate the output equation of the inverse dynamics of the baseline feedback control system. 
Along this particular test trajectory, the root-mean-square (RMS) modeling errors of the DNN for systems~(\ref{eqn:system_sim_stable}) and (\ref{eqn:system_sim_unstable}) are approximately $2\times 10^{-6}$ and $0.06$, respectively. 
In spite of having good modeling accuracy, one can see from Fig.~\ref{subfig:sim_output} that for the non-minimum phase system~(\ref{eqn:system_sim_unstable}), when the DNN module is pre-cascaded to the system, the output response of the system (blue) suffers from numerical issues.
Indeed, even when Eqn.~(\ref{eqn:underlyingFunction_linSS}) is used to calculate the reference and propagate the system forward in time, due to the inherent instability, the precision of the computation quickly falls and leads to unbounded output. Thus, as reflected in this example, even with an unstable zero very close to the unit circle, the associated numerical issues prevent the DNN module from effectively improving the performance of the closed-loop system.
\com{In contrast, for the minimum phase system~(\ref{eqn:system_sim_stable}), the DNN module approximates the output equation of the inverse dynamics, and the corrected reference signal generated by the DNN module approximately achieves an identity mapping from the desired output (red) to the actual output (blue).} 
The RMS tracking error for this DNN-enhanced system is of the order of $10^{-6}$~m, which is only limited by the modeling accuracies and numerical precision in this simulation setting.

\section{Quadrotor Experiments}
\label{sec:experiment}
In this section, \textit{Insight~2} and \textit{Insight~3} are illustrated using experiments on a quadrotor vehicle. 

\subsection{Experiments Setup}
\label{subsec:expSetup}
\edit{The full state of the quadrotor consists of the translational positions $\mathbf{p}=\left(x,y,z\right)$ and velocities $\mathbf{v}=\left(\dot{x},\dot{y},\dot{z}\right)$, the roll-pitch-yaw Euler angles $\boldsymbol{\theta}=\left(\phi,\theta,\psi\right)$, as well as the rotational velocities $\boldsymbol{\omega}=\left(p,q,r\right)$. The control problem is to design a controller such that the translational position of the center of mass of the quadrotor precisely tracks the desired trajectories $x_{d}(t)$, $y_{d}(t)$, $z_{d}(t)$. 
}

\edit{The baseline controller (grey box in Fig.~\ref{fig:blkdiag}) is a standard nonlinear controller consisting of a nonlinear transformation and a PD controller (see~\cite{DNNimpromptuTrack} for more details). In the DNN-enhanced scenario, a DNN module is used to correct the position and velocity reference signals sent to the baseline controller. 
For the experiments, 
the DNNs are fully-connected FNNs with 4 hidden layers of 128 rectified linear units (ReLU). 
The inputs to the DNN module are selected based on \textit{Insight~2} and compared to the selection in \cite{DNNimpromptuTrack}. The outputs of the DNN module are the translational position and velocity references ($\mathbf{p}_r$ and $\mathbf{v}_r$) given to the baseline controller. 
}

\subsection{DNN Feature Selection}
\label{subsec:featureSelection}

\edit{In \cite{DNNimpromptuTrack}, through experimental trial-and-error,
the DNN module input selection that led to efficacious performance was found to be $\mathcal{I}=\{\mathbf{p}_d(t+4)-\mathbf{p_a}(t),\mathbf{p}_d(t+6)-\mathbf{p_a}(t), \mathbf{v}_a(t),\mathbf{v}_d(t+4),\mathbf{v}_d(t+6),\boldsymbol{\theta}_a(t),\boldsymbol{\theta}_d(t+4),\boldsymbol{\theta}_d(t+6),\boldsymbol{\omega}_a(t),\boldsymbol{\omega}_d(t+4),\boldsymbol{\omega}_d(t+6),\ddot{z}_a(t),\ddot{z}_d(t+4),\ddot{z}_d(t+6)\}$, where the subscripts $a$ and $d$ denote the actual and desired values, respectively. In this subsection, we show that by following \textit{Insight~2}, a similar performance as in \cite{DNNimpromptuTrack} can be obtained with significantly less DNN inputs.} 
For the fairness of the comparison, the baseline controller, the DNN architecture, and the training process are identical to \cite{DNNimpromptuTrack}; the only difference is the selected inputs to the DNN module.

\begin{figure}[t]
\centering
\includegraphics[trim={0.5cm 0 1.2cm 0},clip,width = 0.5\textwidth]{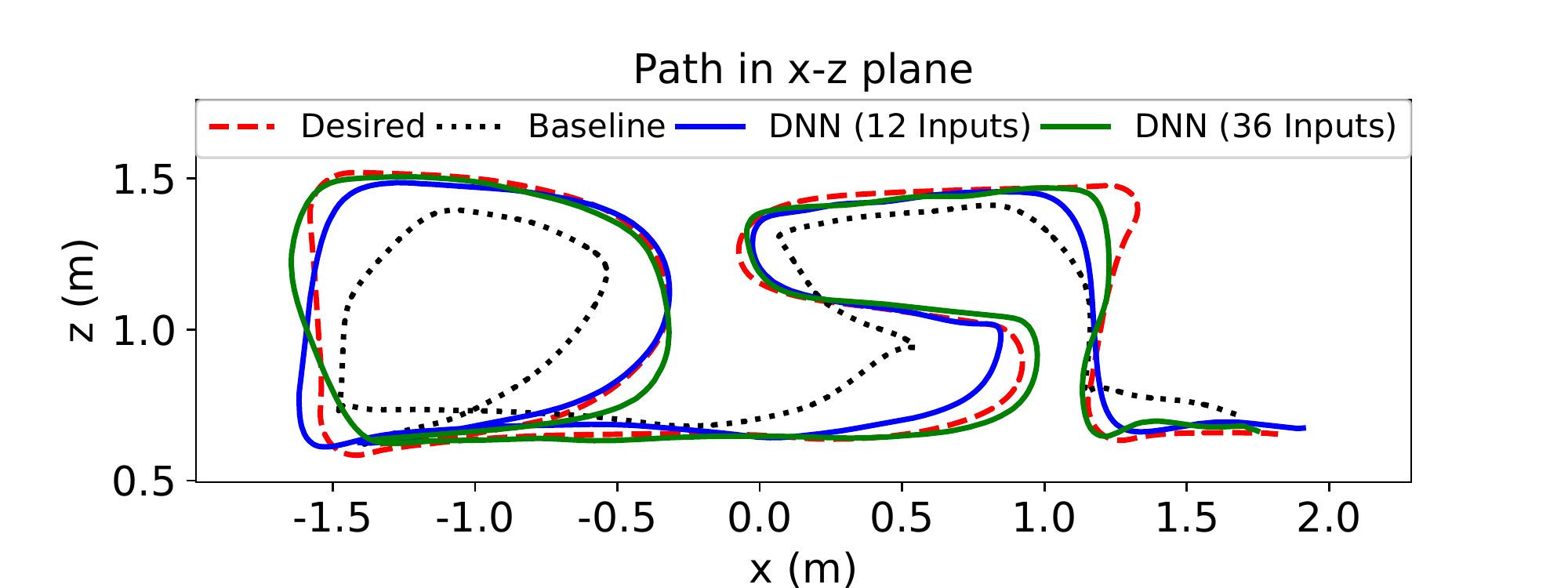}
\includegraphics[trim={0.5cm 0 1.2cm 0},clip,width = 0.5\textwidth]{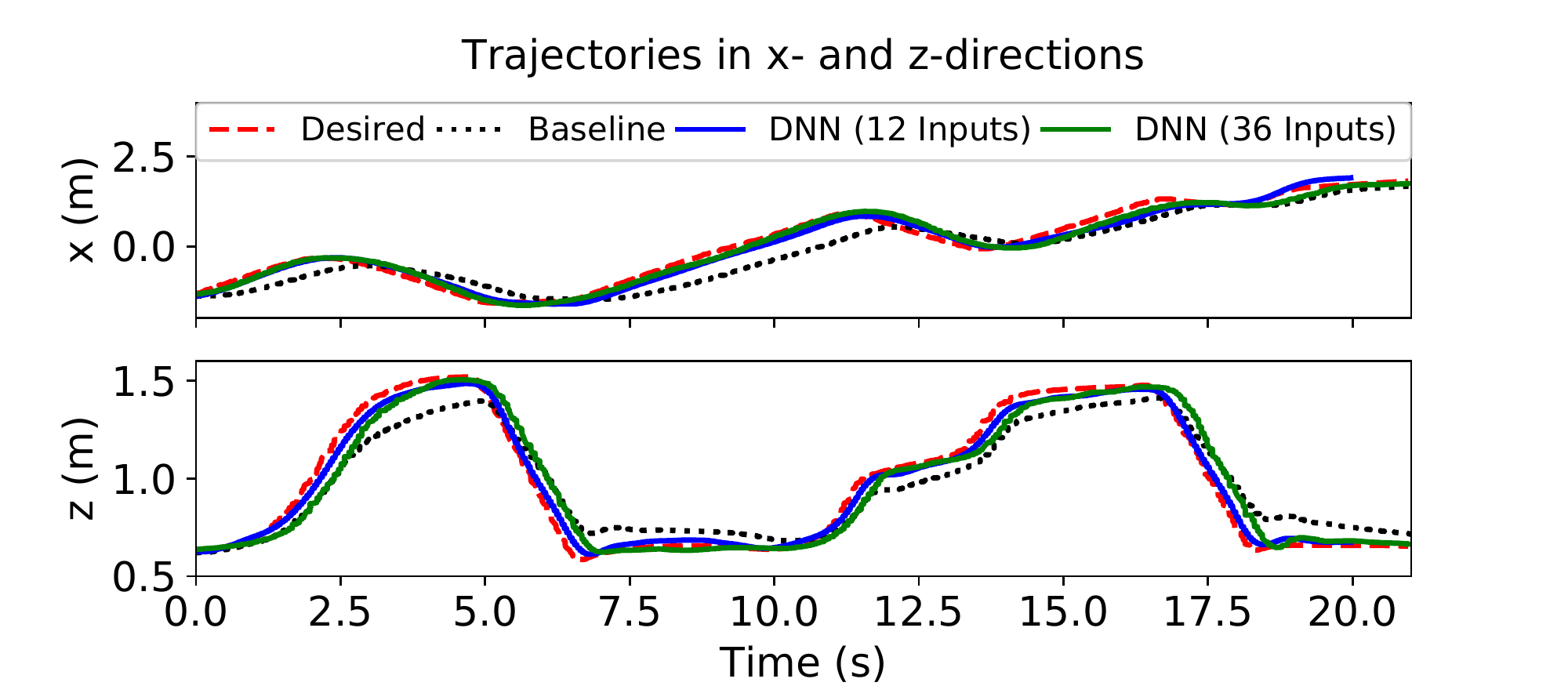}
\vspace{-1.8em}
\caption{Comparison between the DNN-enhanced systems based on the feature selection in \cite{DNNimpromptuTrack} (DNN with 36 inputs) and the one using \textit{Insight~2} (DNN with 12 inputs). From the path \textit{(top)} and the trajectory \textit{(bottom)} plots, despite having only $1/3$ of the inputs, the DNN module designed based on \textit{Insight~2} leads to similar tracking performance improvements.}
\label{fig:exp_dsl}
\vspace{-0em}
\end{figure}

For implementing the state space approach of \textit{Insight~2}, we first examined the step responses of the baseline feedback system, and the relative degrees of the system are determined to be $4$, $4$, and $2$ in the $x$-, $y$-, and $z$-directions, respectively. Moreover, following \textit{Insight~3}, since the step responses of the baseline system have zero steady state errors, we apply the difference learning scheme as in~\cite{DNNimpromptuTrack}. 
\rev{Based on \textit{Insight~2} with the difference learning scheme and by assuming that exact tracking is approximately achieved using the DNN-enhanced architecture, the DNN module inputs are selected to be $\mathcal{I}=\{x_d(t+4)-x_d(t),y_d(t+4)-y_d(t),z_d(t+2)-z_d(t),\dot{x}_d(t+3)-\dot{x}_d(t),\dot{y}_d(t+3)-\dot{y}_d(t),\dot{z}_d(t+1)-\dot{z}_d(t),\boldsymbol{\theta}_a(t),\boldsymbol{\omega}_a(t)\}$. In comparison, the DNN module designed based on \textit{Insight~2} with the difference learning has 12 inputs, while that in~\cite{DNNimpromptuTrack} has 36 inputs.} 
Fig.~\ref{fig:exp_dsl} shows the performance of the two DNN modules on a 2D test trajectory. Similar comparisons are carried out on four additional test trajectories; Table~\ref{tab:percentageReduction} summarizes the RMS tracking error reduction achieved by the two DNN-enhanced systems. \edit{From these results, it can be seen that despite having significantly less input features, the DNN trained with features selected based on \textit{Insight~2} leads to comparable performance to that in~\cite{DNNimpromptuTrack}. 
From the trajectory plots in Fig.~\ref{fig:exp_dsl}, it can be seen that the delay in the $z$-direction is reduced with the new feature selection, where relative degrees are properly identified.}

\vspace{0em}
\begin{table}[!t]
\centering
\caption{\vspace{0em} Percentage Reduction in RMS Tracking Error}
\vspace{-.8em}
\label{tab:percentageReduction}
\footnotesize
\begin{tabular}{c|ccccc|c}
\hline\hline
Traj. ID & `DSL' & 2 & 3 &4 & 5& Avg. \\\hline
12 Inputs & 52.4\%& 48.4\% & 42.0\% &46.6\% &36.9\% &45.2\% \\ 
36 Inputs & 55.6\% &  61.4\% &  39.7\% &  43.9\%&  26.6\%&  45.5\%\\
\hline\hline
\end{tabular}
\vspace{-1em}
\end{table}

\subsection{Difference Learning}
\label{subsec:diffLearning}


\begin{figure}[t]
\centering
\vspace{0.6em}
\includegraphics[trim={1.2cm 0 1.2cm 0},clip,width = 0.5\textwidth]{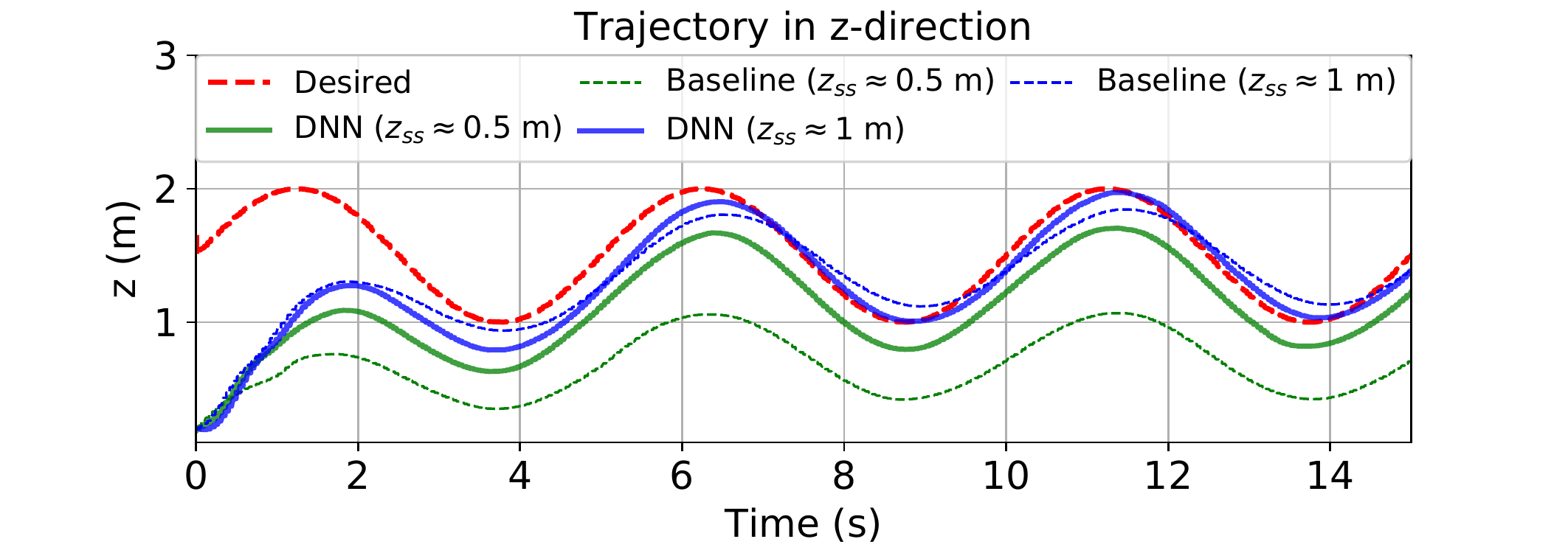}
\vspace{-2em}
\caption{Comparison of applying the difference learning scheme to the DNN for the two baseline systems -- one achieves zero steady state error for step references, which is the necessary condition specified in \textit{Insight~3},  and the other one does not. When this necessary condition is not achieved, the application of the difference learning scheme prevents the DNN module from properly compensating for biases in the response.}
\label{fig:learningDiff}
\vspace{-1em}
\end{figure}

\edit{In Section~\ref{subsec:featureSelection}, we showed that when the baseline system satisfies the condition specified in \textit{Insight~3}, the DNN module trained with the difference learning scheme effectively improved the tracking performance. 
In this subsection, in order to illustrate the necessity of the condition in \textit{Insight~3}, the same input features, DNN architecture, and training process are applied to a modified feedback control system, namely the same baseline system with a factor of 0.5 multiplied to the reference input $z_r$. 
For this modified baseline system, the steady state value of the output in the $z$-direction, denoted $z_{ss}$, is approximately 0.5~m for a unit step reference. Fig.~\ref{fig:learningDiff} shows the performance of the DNN-enhanced system for the original and modified baseline system scenarios. For the original scenario (blue), the DNN module trained with the difference learning scheme results in good tracking performance; however, for the modified scenario (green), the DNN module cannot fully compensate for the bias in $z$. This result is consistent with the discussion in Section~\ref{subsec:differenceLearning}.}





\section{Conclusions and Future Work}
\label{sec:conclusions}
We provided theoretical insights into a DNN-enhanced control architecture for achieving high-accuracy, impromptu tracking. These insights represent general design guidelines for applying the DNN-enhanced architecture to any practical system. 
Through theoretical derivations, simulations and experiments, we showed that the DNN module in the proposed architecture is an approximation of the output equation of the inverse dynamics of the baseline system. 
Due to this association, we illustrated that the DNN-enhanced control architecture (initially proposed in \cite{DNNimpromptuTrack}) may not be effective for closed-loop systems with unstable zero dynamics. 
We also provided guidelines for efficiently selecting the DNN features, which led to a performance similar to previous trial-and-error techniques \cite{DNNimpromptuTrack} but had a significantly lower DNN input dimension. 
Moreover, it is shown through theory and experiments that the applicability of the difference learning scheme in \cite{DNNimpromptuTrack} relies on a necessary condition: the baseline system achieves zero steady state errors for step references. 

\com{Potential extensions of this work include the exploration of approaches for adapting the proposed DNN-enhanced architecture to non-minimum phase systems, and the incorporation of uncertainty estimations in the DNN-based learning approach.
}

\bibliographystyle{IEEEtran}
\bibliography{IEEEabrv,reference}

\end{document}